%% file: KivaCI.tex
\documentclass[12pt]{article}
\usepackage{datetime}
\newdateformat{monthyeardate}{
	\monthname[\THEMONTH], \THEYEAR}
  
\usepackage{framed}

\usepackage{array}
\newcolumntype{L}[1]{>{\raggedright\let\newline\\\arraybackslash\hspace{0pt}}m{#1}}
\newcolumntype{C}[1]{>{\centering\let\newline\\\arraybackslash\hspace{0pt}}m{#1}}
\newcolumntype{R}[1]{>{\raggedleft\let\newline\\\arraybackslash\hspace{0pt}}m{#1}}
  
\usepackage[table]{xcolor}
\usepackage{booktabs}
\usepackage{tabularx}
\usepackage{tcolorbox}

\usepackage{csquotes}
  
\usepackage{mathtools}
 \DeclarePairedDelimiterX{\norm}[1]{\lVert}{\rVert}{#1}
\usepackage{algorithm}
\usepackage{relsize}
\usepackage{algpseudocode}
\usepackage{pifont}
\usepackage{latexsym}
\usepackage{amsmath}
\usepackage{amsfonts}
\usepackage{amssymb}
\usepackage{amsthm}
\usepackage{multirow}
\usepackage{setspace}
\usepackage{graphicx}
\usepackage{caption}
\usepackage{subcaption} 
\usepackage[titletoc,title]{appendix}
\usepackage{afterpage}
\usepackage{xcolor, colortbl} % color table rows/cols
\definecolor{LightCyan}{rgb}{0.88 , 1, 1}
\definecolor{Gray}{gray}{0.85}
\usepackage{booktabs} % make nice tables

\usepackage{multirow, multicol}
\usepackage{epsfig, subfloat}
\usepackage{pstricks, sgamevar, egameps}
\usepackage{url}

\usepackage{breakurl}
\usepackage[breaklinks]{hyperref}

\usepackage{verbatim}

\usepackage{pgfplots}
\usepackage{tikz}
\usepackage{tikz-dependency}
\usetikzlibrary{positioning,chains,fit,shapes,arrows,calc}
\usepackage{tkz-graph}
\usetikzlibrary{decorations.text}
\usetikzlibrary{decorations.pathmorphing}
\usetikzlibrary{decorations.markings}

\usepackage{geometry}
\newgeometry{margin=1in}

\newcommand\beq{\begin{equation}}
\newcommand\eeq{\end{equation}}
\newcommand\bit{\begin{itemize}}
\newcommand\eit{\end{itemize}}
\newcommand\bea{\begin{eqnarray}}
\newcommand\eea{\end{eqnarray}}
\newcommand\beas{\begin{eqnarray*}}
\newcommand\eeas{\end{eqnarray*}}
\newcommand\beqa{\begin{eqnarray}}
\newcommand\eeqa{\end{eqnarray}}

\newtheorem{theorem}{Theorem}[section]

\newtheorem{assumption}[theorem]{Assumption}
\theoremstyle{remark}

\numberwithin{equation}{section}

\theoremstyle{remark}

\begin{document}
\title{A Deep Causal Inference Approach to Measuring the Effects of Forming Group Loans in Online Non-profit Microfinance Platform\thanks{We are grateful to Tim Dozat for many valuable suggestions. We also thank Mohsen Bayati, Dan Cao, Gabriel Carroll, Han Hong, and Quoc Le for helpful discussions. All errors are ours.}}
\author{
  Thai T. Pham\footnote{Graduate School of Business, Stanford University. Email: \texttt{thaipham@stanford.edu}. Thai Pham thanks his advisor Guido Imbens for supporting him throughout the project.}
\and
  Yuanyuan Shen\footnote{Graduate School of Business, Stanford University. Email: \texttt{yyshen@stanford.edu}.}}

\date{\monthyeardate\today}
\maketitle

\sloppy % avoids the breakage of words at the end of lines

\begin{abstract}
\input{abstract}
\end{abstract}

\newpage
\input{intro}

\input{literature}

\input{data}

\input{causal}

\input{pre_analysis}

\input{method}

\input{whyDL}

\input{results}

\input{conclusion}

\input{reference}

\newpage
\input{appendix}

% ------------- & ------------- % -------------- & ------------- % ------------- & ------------- %
%%%%%%%%%%%%%%%%%%%%%%%%%%%%%%%%%
%%%%%%%%%%%%%%%%%%%%%%%%%%%%%%%%%
%%%%%%%%%%%%%%%%%%%%%%%%%%%%%%%%%
%%%%%%%%%%%%%%%%%%%%%%%%%%%%%%%%%
%%%%%%%%%%%%%%%%%%%%%%%%%%%%%%%%%
%%%%%%%%%%%%%%%%%%%%%%%%%%%%%%%%%
\end{document}

%% file: abstract.tex
 Kiva is an online non-profit crowdsouring microfinance platform that raises funds for the poor in the third world. The borrowers on Kiva are small business owners and individuals in urgent need of money. To raise funds as fast as possible, they have the option to form groups and post loan requests in the name of their groups. While it is generally believed that group loans pose less risk for investors than individual loans do, we study whether this is the case in a philanthropic online marketplace. In particular, we measure the effect of group loans on funding time while controlling for the loan sizes and other factors. Because loan descriptions (in the form of texts) play an important role in lenders' decision process on Kiva, we make use of this information through deep learning in natural language processing. In this aspect, this is the first paper that uses one of the most advanced deep learning techniques to deal with unstructured data in a way that can take advantage of its superior prediction power to answer causal questions. We find that on average, forming group loans speeds up the funding time by about 3.3 days. \\

\textit{Key words:} group loan, treatment effect, philanthropic, online, crowdfunding, microfinance, deep learning.

%% file: intro.tex
\section{Introduction}
\label{sec:intro}

Poverty has long been a great concern for our world. To alleviate this problem, microfinance emerged with the initiation of the Nobel Peace Prize-winning Grameen Bank in Bangladesh in 1976. Since then, many microfinance institutes (MFIs) were established in the third world, extending loans with moderate interest rates to the villagers who lack collaterals to secure their loans. Financing is one of the major challenges these local MFIs have been facing. They need a substantial amount of capital to lend to the poor as well as bear the high default risk. Fortunately, the arrival of the digital age makes it much easier for them to raise funds to help the poor. In particular, online crowdsourcing platforms allows multiple individuals to contribute to a loan that could be life-saving for a poor in need in the third world. 

Kiva is such an online non-profit peer-to-peer microlending platform that connects borrowers with the lenders from all over the world. The Kiva website lists thousands of borrowers who are looking for loans to grow their businesses, go to school, or use for other major events in their daily lives. The lenders on the Kiva website are non-profit seeking and philanthropic. They first browse through the profiles of different borrowers and decide whom they will lend to. They can choose whatever amount to lend (in increments of \$25 USD). After a loan is fully funded, the borrowers pay back according to a pre-determined payment schedule. If insufficient funds are collected by the expiration date, all the lenders for that loan will be refunded; otherwise, after the loans are fully repaid, the lenders get their original amount of money back from the partners with no interest. In case of a default, it is usually the lenders that bear the loss. Nevertheless, Kiva maintains a default rate as low as $4.8\%$. 

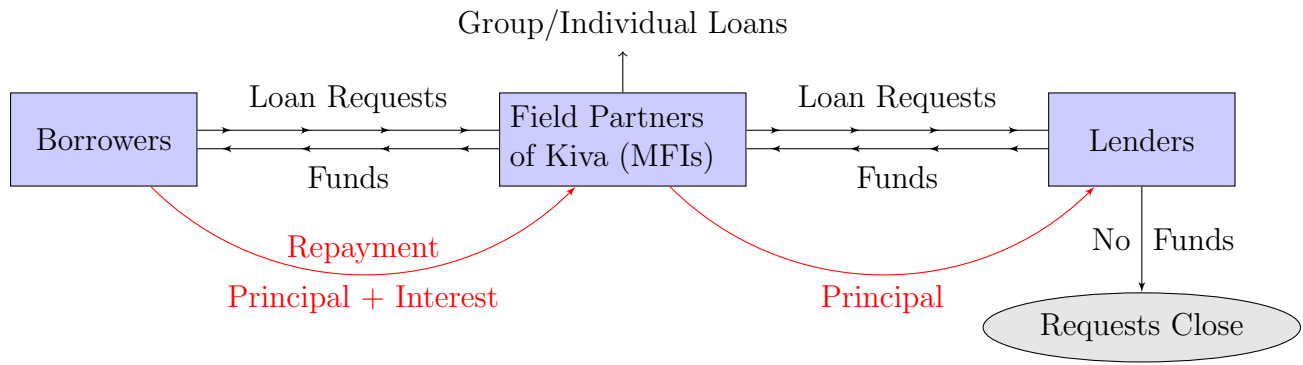
\begin{figure}[h]
\begin{center}

\tikzstyle{block} = [draw, fill=blue!20, rectangle, 
    minimum height=3em, minimum width=6em]
\tikzstyle{extension} = [draw, fill=gray!20, ellipse,
    minimum height=2em, minimum width=5em]
\tikzstyle{pinstyle} = [pin edge={-to,thin,black}]

\tikzstyle{arrowed double line}=[
            double distance between line centers=7pt,
            postaction=decorate,
            decoration={
                markings,
                mark=between positions 10pt and -10pt step 30pt with {
                    \arrow[thin,yshift=3.5pt,
                     xshift=3pt]{>}
                    \arrow[thin,yshift=-3.5pt,
                     xshift=3pt]{<}
                },
            },
        ]

\begin{tikzpicture}[auto, node distance=3cm,>=latex']
    % Draw nodes
    \node [block] (B) {Borrowers};
    \node [block, right of=B,
    pin={[pinstyle]above:Group/Individual Loans},
            node distance=6.9cm, text width=3cm] (F) {Field Partners of Kiva (MFIs)};
    \node [block, right of=F,
            node distance=6.9cm] (L) {Lenders};
            
    \node [extension, below of=L,
            node distance=2.5cm] (P) {Requests Close};
    
    % Draw edges
    \draw [-] (B) -- node[above=0.5em] {Loan Requests} node[below=0.5em] {Funds} (F);
    \draw [-] (F) -- node[above=0.5em] {Loan Requests} node[below=0.5em] {Funds} (L);
    \draw [->] (L) -- node[left=0.005em] {No} node[right=0.005em] {Funds} (P);

    \draw [red, ->] (B) to[out=-45,in=-135] node[above] {Repayment} node[below] {Principal + Interest} (F);
    \draw [red, ->] (F) to[out=-45,in=-135] node[below] {Principal} (L);
    
    \draw[arrowed double line] (B) to (F);
    \draw[arrowed double line] (F) to (L);
    
\end{tikzpicture}
\caption{Diagram of Borrowers, Kiva's Field Partners (MFIs), and Lenders.} 
\label{Fig:Kiva_Diagram}
\end{center}
\end{figure}

Kiva relies on the MFIs in the third-world, the so-called Field Partners, to screen the borrowers, post loan requests on the website, disburse loans, and collect repayments. The field partners have wider exposure to the poor, who are in general not readily accessible to the Internet. They serve as bridges to connect borrowers with lenders on Kiva. More often, they pre-disburse the loans to the poor in need even before the loans are fully funded on Kiva. Hence, their goals are to raise funds as fast as possible so that they can pre-disburse the amount to other people in need of money. The speed of funding is critical for their financing. From Kiva's side, a considerable speed of funding is also a strong sign on its competitiveness in the microlending market, which is in align with the borrowers' goals. The relationship among the borrowers, the field partners, and the lenders is visualized in Figure \ref{Fig:Kiva_Diagram}.

To raise funds as fast as possible, the field partners may choose to suggest individual borrowers form into groups to raise their credibility to the lenders. Group lending (or joint liability) was first introduced in Bangladesh to mitigate the adverse selection problem (Armendariz and Morduch \cite{AM2007}). Under joint liability, all group members are responsible for the loans of other group members. If one member defaults, the other group members are required to cover the loss. Hence, it is in everyone's interest to ensure that the other group members pay. The traditional MFIs encourage the borrowers to form groups and apply for loans in the name of groups to increase their chance of getting approved because they believe that safe borrowers who know each other are likely to form their own groups (Hossain \cite{H1998}, Kodongo and Kendi \cite{KK2013}). This pooling of borrowers also significantly reduces the lenders' risk exposure as if they are investing in a variety of financial assets. Ghatak \cite{G1999} concludes that group lending increases repayment rates. Similarly, Islam \cite{I1995} suggests that at the same interest rate, the expected rate of repayment is higher with lower risk for group loans. 

While conventional evidence suggests by lending to groups, lenders may get their repayments faster and more reliably,  it may not be the case for Kiva. This is because the lenders on Kiva are non-profit seeking. They can lend whatever amount they want ($\$25$ minimum) and can always choose to spread the risk by making little investments on multiple loans.

In this paper, we focus on answering the main question for the field partners: should they organize the borrowers into groups on Kiva in order for the projects to get funded as quickly as possible? Specifically, we measure the effects of forming group loans on time till the projects get funded using Kiva data. We are mainly interested in the (population) average treatment effects. 

Our results show a significant negative average treatment effect of group lending on the funding time. The most cutting-edge methods give similar estimates of roughly $-3.3$ with small standard deviation of $0.167$. With about $3.3$ days faster on average in terms of funding time when group loan is used, the field partners would want to encourage the borrowers to form groups in requesting loans in general. This 3.3 day period is significant in Kiva's perspective in showing its competitiveness in the microfunding market. 

As a significant part of the data is textual, it is very hard to use traditional econometric methods to deal with it while discarding all such data means that we would leave out a rich source of information. Fortunately, we can take advantage of the new advancement in machine learning literature, namely deep learning, to deal with text data. There is a small recent trend in using machine learning generally and deep learning particularly in answering economic questions. In most of these cases, they are used to deal with high dimensional and sometimes unstructured data such as images and texts. Traditionally, machine learning is used solely for prediction tasks. Recently, researchers combine machine learning with causal inference framework to answer causal questions. In this paper, we combine deep learning with causal inference framework to answer the questions stated above. We name this approach deep causal inference. This paper is the first that uses one of the most advanced deep learning techniques to deal with unstructured data in a way that can take advantage of its superior prediction power to answer causal questions.

The rest of the paper is organized as follows. Section \ref{sec:lit} reviews related literature which includes work about traditional microfinance, the online crowdfunding microfinance, and deep learning for unstructured data in economics. Section \ref{sec:data} discusses and summarizes the Kiva data used in the analysis. Section \ref{sec:causal} covers the causal inference setting. Section \ref{sec:pre_analysis} gives preliminary analysis on the data and the treatment effects. Section \ref{sec:method} focuses on the methods used to answer causal questions: the pre-processing step, the baseline method, and the deep causal inference approach. Section \ref{sec:DLTechniques} explains the deep learning techniques used in this paper. Section \ref{sec:results} discusses results estimated by the considered models. Section \ref{sec:conclusion} concludes. 

%% file: literature.tex
\section{Literature Review}
\label{sec:lit}

\subsection{Online Crowdfunding Microfinance}
There is a small but growing literature in online microfinance platforms.  Various crowdfunding platforms differ in what the backers expect to receive in exchange for their money pledged. Reward-based platforms such as Kickstarter and Indiegogo consist of projects that involve the pre-launch of creative business ideas where lenders can get tangible rewards. Mollick \cite{Mollick} and Qiu \cite{Qiu} have studied the factors that can lead to the success or failure of funding projects on Kickstarter.  
Other Internet-based peer-to-peer lending platforms such as Prosper and Lending Club take the more traditional form where borrowers are expected to pay the original principal as well as a fixed interest rate. Research on this type of platforms (such as Zhang and Liu \cite{ZhangLiu}) tracks the funding dynamics on Prosper and finds that lenders tend to act rationally when the borrower exhibits signals of low quality. 

Kiva differs from these online marketplaces in that its investors (lenders) are non-profit driven. Hence, it is of our interest to study what influences lenders' responsiveness in this setting. Allison et al. \cite{Allison} find that entrepreneur's narrative, i.e. description of a loan, makes a difference. Lenders on Kiva tend to act positively to narratives that frames the business as helping others, and less positively to the narrative that emphasizes on the business opportunity. However, they restrict their attention to loans for businesses. In contrast, we study the effects of group borrowing on funding time for all types of loans. Moreover, they do not use the machine learning approach as we do.

Group loans have been shown to have advantage over individual loans when it comes to microfinance. Several papers in the literature try to explain the mechanism behind this fact. Stiglitz \cite{S1990} and Banerjee et al. \cite{BBG1994} emphasize the moral hazard problems which joint liability lending and monitoring can mitigate. Besley and Coate \cite{BC1995} cares more about the nature of the contract with limited or no enforcement and mostly no collateral requirement. On the other hand, Ghatak \cite{G1999} focuses on using joint liability contracts to overcome the adverse selection problem of borrowers.

In short, the main problems of moral hazard, limited commitment, and adverse selection of borrowers, which prevent them from receiving microfunding, are shown theoretically to be solved by using a group instead of individual loan. 

Using Thai repayment data, Ahlin and Townsend \cite{AT2007} test the predictions of the four afore-mentioned theoretical models. Based on these models they generate theoretical predictions regarding the determinants of the repayment performance of groups.

On the similar line of work, Wydick \cite{W1999} uses borrowing group data from Guatemala to empirically test the effects of peer monitoring via group pressure and social ties on group performance. Paxton et al. \cite{PGT2000} present a stylized model of group loan repayment based on a two-stage econometric model together with an empirical analysis using data from a survey of 140 lending groups in Burkina Faso; they find that group dynamics, as well as other factors, can be important determinants of loan repayment. For an overview of the promise of traditional microfinance, the reader is referred to Morduch \cite{M1999}. 

Our work is also empirical, but we are different in that we do not focus on the repayment activity of the borrowers, instead on the time till their projects get funded. Moreover, we focus on online crowd-funding instead of the traditional system of MFIs.

\subsection{Causal Inference \& Deep Learning in Economics}
Estimating average treatment effects (ATEs) is a canonical problem across many different fields. With the recent rise of big, non-experimental data, researchers have shifted attention to estimating treatment effects in observational data of high dimensions. Athey et al. \cite{AIPW2017} and Imbens and Pham \cite{IP2017} review a list of estimators in this setting. Among these methods, we particularly pay attention to the Double Selection Estimator (DSE) by Belloni et al. \cite{BCH2014}, the Doubly Robust Estimator (DRE) with the main ideas dated back to Robins et al. \cite{RRZ1994}, and Targeted Maximum Likelihood Estimator (TMLE) by Van Der Laan and Rubin \cite{VR2006}. These methods will be reviewed in detail in Section \ref{subsec:advance}. In short, we use DSE because it is simple to implement and it works reasonably in many cases. We use DRE and TMLE because they have the important double robustness property and they can incorporate effectively the advanced machine learning and deep learning techniques into the models. 

There is a line of work that uses deep learning and related natural language and imagery processing tools to deal with unstructured data in economics as well as in social sciences in general. Gentzkow et al. \cite{GST2015} use text processing tools on speech data to estimate political affiliations. Also using text data, Kang et al. \cite{KKLC2013} forecast restaurant hygiene levels with Yelp reviews. Differently, Jean et al. \cite{JBXDLE2016} attack the poverty problem with satellite images and deep learning methods. Similarly, Naik et al. \cite{NRH2016} use street-view images to investigate the impact of urban appearance on residents' socioeconomic status. Sirignano et al. \cite{SSG2016}, on the other hand, use deep learning for studying mortgage risk. Also using deep learning, though not applying on text or imagery data, Hartford et al. \cite{HLLT2017} propose the Deep Instrumental Variables model to answer causal questions when endogeneity is present. 

We also need to deal with a lot of textual data, which comes from loan descriptions of the borrowers on Kiva. This is exactly where the deep learning approach can come into play.  

%% file: data.tex
\section{Data}
\label{sec:data}

Our cleaned data set has $995,911$ loan entries from Kiva. We obtain the raw data from the Kiva website \url{http://build.kiva.org/} from Jan $1^{st}$, 2006 to May $10^{th}$, 2016. We describe the data below. 

The outcome of interest is the time till the borrowers get funded. Since 95.2\% of the loans posted on Kiva are fully funded. we focus our attention on these loans. Among them the fastest funding time is $15$ seconds, the longest funding time is $154$ days, which is roughly $5$ months. The average funding time is $7.11$ days and the median funding time is $1.79$ days.

The treatment we care about in this study is the decision whether to form a group or not in requesting loan. We are interested in estimating the effects of group borrowing relative to individual borrowing on the funding time. We want to answer this question in the context of philanthropic online crowd lending instead of the traditional MFIs. Both the philanthropic and the online crowdsourcing aspects make our question different from all those in the literature regarding the role of group loans.  

In our data, about $14.3\%$ of the loans are from group borrowers. It is thus interesting to see if forming a group would help getting funded  or getting funded faster. If it would, then it would change the way people apply for loans.  

After preprocessing, the covariates (features) of each observation (i.e., loan) include a description about the situation of the borrowers together with the reason why they are looking for funding, and other information. The raw data, which is presented in the Appendix, contains more information than what we use in this study. However, we do not include the data that is not informative, hard to use, or unrelated to our study. We refer the reader to Appendix for more details. A list of the covariates used are summarized in Table \ref{tab:covVar}. 
    
\begin{table}[H]
\caption{Covariate Variable Summary}
\label{tab:covVar}
\begin{center}
\begin{tabular}{lll}
\toprule
\multicolumn{2}{c}{Item} \\
\cmidrule(r){1-2}
Variable       & Description                    & Type \\
\midrule
\texttt{description\char`_texts}           & Loan description     &   Text     \\
\texttt{loan\char`_amount}  & The amount borrower requests  & Numerical      \\
\texttt{sector}    & Purpose of the loan     &  Categorical\\
\texttt{risker} & Lenders or partners bear the default risk & Binary \\
\texttt{gender} & The gender of the borrower(s) & Binary \\ 
\bottomrule
\end{tabular}
\end{center} 
\end{table}

Among the listed variables, we notice the important text data \texttt{description\char`_texts}. One example of the data is as follows; this is an individual loan. 

\begin{displayquote}
    \textit{
    ``James is a 35-year-old mixed crop farmer. He is married to Zipporah, a housewife. They are blessed with two children age seven and four years old, respectively. James has been practicing farming for the past two years with a monthly income of KES 18,000. James is applying for his third loan from KADET LTD after repaying the previous loans successfully. He will use the loan to buy poultry feed and one-day-old chicks for rearing. With the anticipated profit from the business, he will expand his poultry farming. His hopes are to buy a car and venture into the transport business.''}
\end{displayquote}
    
Another example is a group loan:
\begin{displayquote}
    \textit{
    ``Aruna, age 35, is married with 3 children (ages 12, 9, and 5). She has a business selling timber that she started three years ago. She works from 9am to 3pm daily and now makes a monthly profit of about $\$280$. Aruna hopes to get a loan in order to increase her stock of timber. She will share this loan with her subgroup, who have businesses dealing in clothing, livestock, food, and firewood sales.''}
\end{displayquote}
    
This text data is hard to deal with using traditional econometric methods. The text descriptions could greatly vary in terms of words used depending on the content of the request. This poses a real challenge to social science researchers who want to utilize this data to answer interesting causal questions.  
    
In a different concern, the \texttt{sector} variable has $15$ different categories. We then replace it with $14$ dummy variables corresponding to the categories. So beside the text data, we have $17$ covariates. Now, we look at more statistics for some of the described variables.

\begin{figure}[H]
    \centering
    \includegraphics[width = 0.9\textwidth]{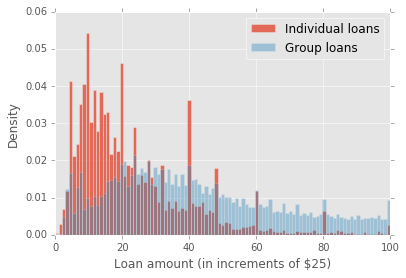}
    \caption{The distribution of loan amounts in increments of \$25.}
    \label{fig:loan_amount}
\end{figure}

Figure \ref{fig:loan_amount} shows the distribution of loan amounts in increments of \$25 dollars.\footnote{On Kiva, both loan amount (for borrowers) and funding  amount (for lenders) are in increments of \$25.} The average loan amount for individual loans is \$652 and that for group loans is \$1816. 

\begin{figure}[H]
    \centering
    \includegraphics[width = 0.9\textwidth]{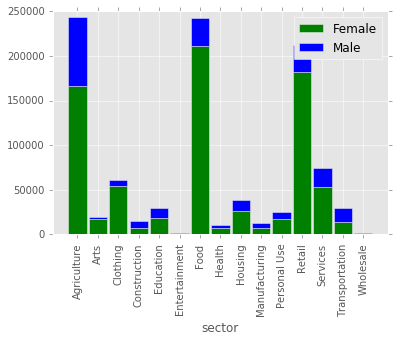}
    \caption{Number of loans in each sector for each gender.}
    \label{fig:num_loan_sec}
\end{figure}

Figure \ref{fig:num_loan_sec} shows the number of loans in each category and the gender (or, in groups' case, the majority of the genders) of the borrower(s).  Most of the borrowers are female and are fueling their communities with food at markets, small grocery stores and restaurants.

\begin{figure}[H]
    \centering
    \includegraphics[width = 0.9\textwidth]{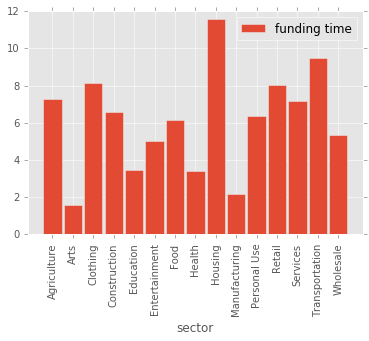}
    \caption{Average funding time in each sector.}
    \label{fig:funding_time_sector}
\end{figure}

What types of loans get funded more quickly? On average, housing takes the longest ($11.55$ days) to be fully funded. Transportation and clothing take the second longest ($9.55$ and $8.19$ respectively). On the other hand, requests on loans for arts, manufacturing, health, and education take the shortest time ($1.57, 2.20, 3.47$, and $3.49$ days respectively) to fulfill on average. See Figure \ref{fig:funding_time_sector} for more details.

\begin{figure}[H]
    \centering
    \includegraphics[width = 0.8\textwidth]{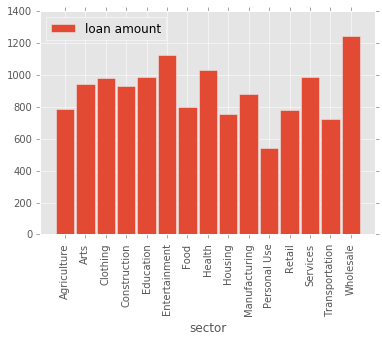}
    \caption{Average loan amount in each sector.}
    \label{fig:loan_amount_sec}
\end{figure}

What types of loans request more financial help? We look at the average loan amount for each sector. As shown in Figure \ref{fig:loan_amount_sec}, wholesale loans are of the largest amounts ($\$1,228$) on average while loans related to entertainment and health correspond to the second and third largest amounts ($\$1,105$ and $\$1,027$ respectively). Since health is also among the sectors that are fulfilled the fastest, we may conclude that health loans are more attractive to lenders in general. On the other hand, loans for personal use are less attractive. Although borrowers of this type request for the least amount ($\$541$ on average), they need a long time to get fully funded.

%% file: causal.tex
\section{Causal Inference Setting}
\label{sec:causal}

We adopt the Rubin Causal Model or the Potential Outcome framework to estimate treatment effects. We start with denoting by $X$ the covariate vector (which includes loan description and other information), $W$ the binary treatment variable (decision whether to form group loan or not), and $Y$ the outcome of interest (time till the project gets funded). Assume the dataset we obtain is $(Y_i, W_i, X_i)_{i = 1}^n$. Let $n_t = \sum_{i = 1}^n W_i$ and $n_c = n - n_t$. We are interested in estimating the average treatment effects (ATE) of $W$ on $Y$. Specifically, we want to estimate $\tau$, where
    $$\tau = \mathbb{E} [Y(1) - Y(0)].$$
Here $Y(1), Y(0)$ are potential outcomes with exactly one of which observed for each unit. 

In order to estimate $\tau$, we need to make several assumptions, which are all common in the causal inference literature (see Imbens and Rubin \cite{IR2015} for an overview). We restate them here with discussion for why the assumptions may likely hold. 

\begin{assumption}
\label{ass:SUTVA} (SUTVA) One's outcome is not affected by others' treatment decisions (no interference) and the value of the treatment is the same across treated individuals (no variation in treatment value).
\end{assumption}

The treatment variable in this case is the decision whether to form the group loan, so its value is certainly constant across treated units. Moreover, the decision whether to form the group loan depends solely on each project (i.e., each unit); hence, the outcome for each project or the funding time is independent of treatment decisions on other projects. 

Some people may argue that lenders have limited financial assets so funding one project means they have less money to fund others; this may create the interference effect. In Kiva setting, however, lenders can lend any amount of money they want (in the increment of $\$25$) and the amount requested on each loan is small; thus, we can confidently assume away this issue. 

\begin{assumption}
\label{ass:unconfoundedness} (Unconfoundedness) Conditional on observed covariates, the potential outcomes are independent of treatment: 
\begin{equation*}
	Y(0), Y(1) \perp W | X.
\end{equation*}
\end{assumption}
This assumption, also called exogeneity, means that there is no unobserved covariate that could simultaneously affect $W$ and $Y$. This assumption allows one to attribute the cause of the effects $Y(1) - Y(0)$ to only treatment $W$. While being easy to understand, this assumption is almost impossible to validate. 

In our setting, the borrowers tend to form groups for the sake of convenience, e.g. closely related people, rather than being forward looking to the potential outcomes; this means the treatment assignment process can be considered exogenous. There might exist some endogeneity, though this endogeneity should be small enough to be ignored. If this endogeneity is large, then everyone would think forming group loans will speed up the funding process and everyone will tend to do so (following the practice in traditional microfinance); we know that this is not the case here. Hence, we can assume that unconfoundedness holds.  

\begin{assumption}
\label{ass:overlap} (Overlap) For all $X$, the following holds with probability one:
\begin{equation*}
    0 < \mathbb{P} (W = 1 | X) < 1.
\end{equation*}
\end{assumption}
This assumption means there are always observations in each treatment group. 

With these assumptions, we can estimate the treatment effects. 

%% file: pre_analysis.tex
\section{Preliminary Analysis}
\label{sec:pre_analysis}

In this section, we give some preliminary analysis on the treatment effect of $W$ on $Y$ and provide rationales for our use of independent variables. Overall, the average funding time for the treated group is $7.25$ days and that for the control group is $7.08$ days. The estimated ATE by this naive method is $0.17$ with a standard deviation of $0.027$. % ($\sqrt{var(Y(W = 1)) / (n_t - 1) + var(Y(W=0)) / (n_c - 1)} = 0.027$)

\begin{figure}[H]
    \centering
    \includegraphics[width = 0.7\textwidth]{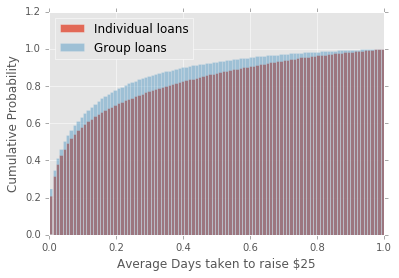}
    \caption{Cumulative probability distribution of the $\frac{\text{funding time}}{\text{loan amount in \$25}}$ ratio}
    \label{fig:time_to_amount}
\end{figure}

However, the naive approach fails to take into account the fact that group loans have in general larger loan amounts, according to Section \ref{sec:data}. As a result, one may want to control for the loan amount, as it is correlated with  both $W$ and $Y$. We first calculate the ratio between $Y$ and the actual loan amount in increments of \$25, referred as the average days taken to raise \$25.  Figure \ref{fig:time_to_amount} displays the cumulative probability distributions of this ratio for group and individual loans respectively, from which we can see the ratio for individual loans stochastically dominates that for group loans. The stochastic dominance is a strong sign for a negative treatment effect on funding time. The mean of the ratio for individual loans is 0.31, suggesting that on average it takes more than 7 hours to raise \$25 dollars. In contrast, the mean of the ratio for group loans is 0.15, suggesting less than 4 hours to raise \$25 on average.

We then run a simple linear regression of $Y$ on loan amount (in whole \$dollars) for treated and control groups and report the results in Table \ref{tab:pre_slm}.

\begin{table}[H]
\caption{Simple Linear Regression of $Y$ on loan amount}
\label{tab:pre_slm}
\begin{center}
\begin{tabular}{@{}l C{5cm} C{5cm} C{2cm} @{}}
\toprule
 & \multicolumn{2}{c}{Coefficient} & R-square  \\
\cmidrule(r){2-3} 
\multicolumn{1}{l|}{Group} & Intercept (s.d.) & Loan amount (s.d)  &    \\
\midrule
\multicolumn{1}{l|}{\texttt{Treated}} &   5.74*** (0.037) &  0.0008*** ($1.54\times 10^{-5}$)      & 0.018 \\
\multicolumn{1}{l|}{\texttt{Control}} & 4.53*** (0.014) &    0.0039*** ($1.43\times 10^{-5}$) & 0.077  \\
\bottomrule
\end{tabular}
\end{center} 
\end{table}

For the treated group, the intercept is higher while the effect of loan amount is much lower. The results suggest that the treatment effect of $W$ on $Y$ depends on the loan amount. If the loan amount is greater than \$390, the average treatment effect is negative (or faster funding); otherwise, the ATE is positive (or slower funding). However, the low $R^2$ suggests that the linear model does not fit our data very well. Hence, we need more advanced methods to model the nonlinearities.

\begin{figure}[H]
    \centering
    \includegraphics[width = 0.7\textwidth]{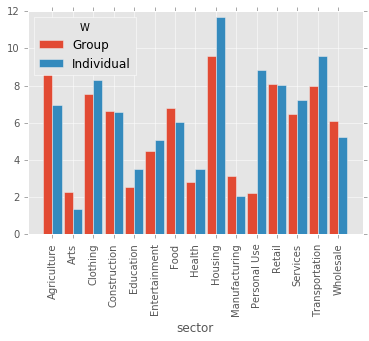}
    \caption{Average funding time in each sector in each treatment group.}
    \label{fig:funding_time_group_sec}
\end{figure}

Figure \ref{fig:funding_time_group_sec} visualizes the average funding time in each sector for each treatment group. Note that we consider only the funded loan requests.

According to this figure, on average the group effects vary across different sectors from negative to neutral to positive. 

\begin{figure}[H]
    \centering
    \begin{subfigure}[b]{0.5\textwidth}
        \centering
        \includegraphics[width = \textwidth]{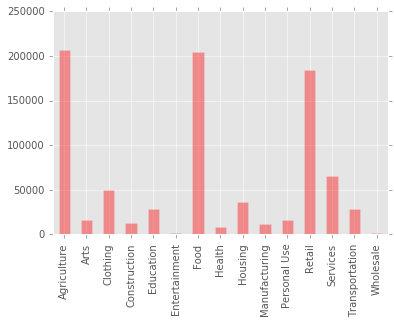}
        \caption{Individual loans}
        \label{fig:count_ind_loan}
    \end{subfigure}%
    \begin{subfigure}[b]{0.5\textwidth}
        \centering
        \includegraphics[width = \textwidth]{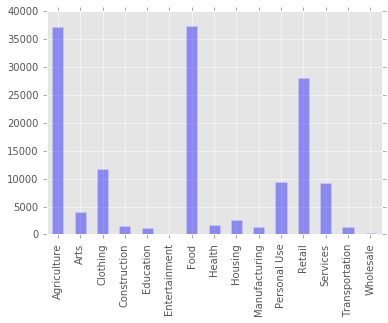}
        \caption{Group loans}
        \label{fig:count_group_loan}
    \end{subfigure}
    \caption{Distribution of the sectors of loans}
    \label{fig:count_sector_loan}
\end{figure}

Figure \ref{fig:count_sector_loan} shows the distributions of loan counts among all sectors for group and individual loans. We can see that food, agriculture and retail are the top three loan categories for both group and individual loans while entertainment and wholesale are the fewest loan categories. Hence, it is highly unlikely that the difference in funding time is due to an uneven distribution of loan types among group and individual loans. 

In the next section, we discuss different methods to estimate this effect more accurately.

%% file: method.tex
\section{Methodology}
\label{sec:method}

Machine Learning has recently been used extensively in the study of causal inference. Researchers have taken advantage of the techniques from this field to estimate both the average treatment effects (Athey et al. \cite{AIW2016}, Belloni et al. \cite{BCFH2013}, Chernozhukov et al. \cite{CCDDH2016}) and the heterogeneous treatment effects (Hill \cite{H2011}, Johansson et al. \cite{JSS2016}, Pham \cite{P2016}, Pham \cite{P2017}, Wager and Athey \cite{WA2016}).

Along with the rise of big data, researchers have access to high-dimensional, structured or unstructured, datasets. Using such data to study causal inference poses an important task but at the same time, an inevitable challenge.

With regard to Kiva's data, the nature of the text data makes it hard for the traditional econometric models to deal well with. However, if we exclude these data, we would lose a lot of information. Fortunately, tools from Machine Learning and especially Deep Learning (Schmidhuber \cite{S2015}) and Natural Language Processing (NLP) (Collobert et al. \cite{CWBKKK2011}) make it possible to utlize this type of data.

\subsection{Preprocessing}
\label{subsec:prepro}

One possible preprocessing step would be to transform the loan description into a numerical vector of fixed dimension; these transformed vectors would then act as a set of covariates to be used in later steps. We can accomplish this transformation step by using the state-of-the-art algorithms in NLP literature: \textit{Word2Vec} or \textit{GloVe} (Mikolov et al. \cite{MSCCD2013} and Pennington et al. \cite{PSM2014}). 

In fact, Word2Vec and GloVe will create word-level vectors. Then depending on the usage, we may combine the vectors of words in each loan description to create a single loan vector. This would give us a high-dimensional covariate vector for each loan description, which can be combined again with other covariates to create the full vector of covariates. (If there is no confusion, we still denote by $X$ this vector or any final-processed covariate vector used in the model.) This step is needed in both the baseline model (discussed in Section \ref{subsec:baseline}) and the advanced ones (see Section \ref{subsec:advance}).

\subsection{Baseline Model}
\label{subsec:baseline}

\subsubsection{Regularized Linear Regression without Text Data}
\label{subsubsec:lrwotext}
In this model, we use only $17$ non-text covariates. We use Linear Regression with elastic-net regularization to estimate two relations: $Y(1) = X \beta_1$ with estimate $\widehat{\beta}_1$ using the treated data and $Y(0) = X \beta_0$ with estimate $\widehat{\beta}_0$ using the control data. Let $\widehat{Y}_1 = X \widehat{\beta}_1$ and $\widehat{Y}_0 = X \widehat{\beta}_0$. The estimator for $\tau$ is 
\begin{equation*}
    \widehat{\tau} = \frac{1}{n} \sum_{i = 1}^n \big( \widehat{Y}_{1, i} - \widehat{Y}_{0, i} \big).
\end{equation*}
To estimate the standard error, we first calculate
\begin{equation*}
    V_1 = \frac{var(Y_i - \widehat{Y}_{1, i} | i: W_i = 1)}{n_t - 1} \text{ and } V_0 = \frac{var(Y_i - \widehat{Y}_{0, i} | i: W_i = 0)}{n_c - 1}.
\end{equation*}
Then, the standard error is estimated by $\sqrt{V_1 + V_0}$.

\subsubsection{Regularized Linear Regression with Text Data}
\label{subsubsec:lrwtext}
In this model, we first pre-process the text data as in Section \ref{subsec:prepro} to create loan vectors. Then we incorporate the loan vectors to $17$ other covariates to create full covariate vectors. Then we proceed as in Section \ref{subsubsec:lrwotext}.  

\subsection{Advanced Model}
\label{subsec:advance}

Athey et al. \cite{AIPW2017} summarize a set of important methods for ATE estimation using Machine Learning in high-dimensional data; we use three of them here: Double Selection Estimator, Doubly Robust Estimator, and Targeted Maximum Likelihood Estimator.

\subsubsection{Double Selection Estimator} 
\label{subsubsec:DSE}

The Double Selection Estimator (DSE) is proposed by Belloni et al. \cite{BCH2014} that uses OLS after variable selection: Use Lasso to select covariates that explain for either treated or control outcomes. Also use Lasso to select covariates that explain for treatment. Take the union of these sets of covariates. Then run OLS for treated and control outcomes separately using the selected covariates. The mean difference on two sets of estimated outcomes on the whole data is the ATE estimate. The estimator for the standard error is similar to that in the baseline model. 

As we can see, this method is easy to implement and is computationally inexpensive. Moreover, it gives reasonably good estimate in many cases. 

\subsubsection{Doubly Robust Estimator} 
\label{subsubsec:DRE}
The second method we use is the Doubly Robust Estimator (DRE) with the main ideas dated back to Robins et al. \cite{RRZ1994} and Robins \cite{R1999}.

We use DRE to take advantage of the deep learning methods in estimating the infinite dimensional components. More importantly, DRE has the double robustness property: it is a consistent estimator of the ATE if either the outcome model or the propensity score model is correctly specified, or both. In high-dimensional setting, it is extremely difficult to guarantee correct specification for both models and thus, this property is essential. The steps of DRE are as follows.

\begin{enumerate}

    \item \textit{Estimating the outcome models}: Use treated data $\{i: W_i = 1\}$ to estimate $\mu(1, x) = \mathbb{E}[Y(1) | X = x]$ with estimator $\widehat{\mu}(1, x)$ and use control data $\{i: W_i = 0\}$ to estimate $\mu(0, x) = \mathbb{E}[Y(0) | X = x]$ with estimator $\widehat{\mu}(0, x)$.
    
    \item \textit{Estimating the propensity score model}: Use all data to estimate $e(x) = \mathbb{P} (W = 1 | X = x)$ with estimator $\widehat{e}(x)$.
    
    \item The DRE $\widehat{\tau}_{DRE}$ for $\tau$ is given by
    \begin{equation*}
        \widehat{\tau}_{DRE} = \frac{1}{n} \sum_{i = 1}^n \left[ W_i \times \frac{Y_i - \widehat{\mu}(1, X_i)}{\widehat{e}(X_i)} - (1 - W_i) \times \frac{Y_i - \widehat{\mu}(0, X_i)}{1 - \widehat{e}(X_i)} + \widehat{\mu}(1, X_i) - \widehat{\mu}(0, X_i) \right].
    \end{equation*}
    
    \item To estimate the standard error, we follow Lunceford and Davidian \cite{LD2004} and use an empirical sandwich estimator. We first define for each $i \in \{1, ..., n\}$, that
\begin{equation*}
	IC_i = W_i \times \frac{Y_i - \widehat{\mu}(1, X_i)}{\widehat{e}(X_i)} - (1 - W_i) \times \frac{Y_i - \widehat{\mu}(0, X_i)}{1 - \widehat{e}(X_i)} + \widehat{\mu}(1, X_i) - \widehat{\mu}(0, X_i) - \widehat{\tau}_{DRE}
\end{equation*}
\begin{equation*}
    \text{and } \sigma^2 = \frac{1}{n} \sum_{i = 1}^n IC_i^2.
\end{equation*}
Then the standard error is estimated by $\frac{\sigma}{\sqrt{n}}$. The $95\%$ Confidence Interval of $\tau$ is estimated by $(\widehat{\tau}_{DRE} - z_{0.975} \frac{\sigma}{\sqrt{n}}, \widehat{\tau}_{DRE} + z_{0.975} \frac{\sigma}{\sqrt{n}})$. 
    
\end{enumerate}

\subsubsection{Targeted Maximum Likelihood Estimator} 
\label{subsubsec:TMLE}
Targeted Maximum Likelihood Estimator (TMLE) can be viewed as a more general version of DRE (Van Der Laan and Rubin \cite{VR2006}). In TMLE, we can also take advantage of the deep learning techniques to estimate the infinite dimensional components. TMLE also has the double robustness property. Moreover as we will see below, TMLE is DRE applied to an updated, better version of the initial infinite dimensional component estimates and thus, TMLE is generally more accurate than DRE (Van Der Laan and Rubin \cite{VR2006}).  

Similar to the first two steps in DRE, we obtain $\widehat{\mu}(1, x), \widehat{\mu}(0, x)$, and $\widehat{e}(x)$. Then define
\begin{equation*}
    H(w, x) = \frac{w}{\widehat{e}(x)} - \frac{1 - w}{1 - \widehat{e}(x)} \text{ and } Q(w, x) = \widehat{\mu}(w, x) + \widehat{\epsilon} \, H(w, x) \text{ for } w \in \{0, 1\},
\end{equation*}
where
\begin{equation*}
    \widehat{\epsilon} = \frac{\sum\limits_{i = 1}^n H(W_i, X_i) (Y_i - \widehat{\mu}(W_i, X_i))}{\sum\limits_{i = 1}^n H(W_i, X_i)^2}.
\end{equation*}
The TMLE $\widehat{\tau}_{TMLE}$ for $\tau$ is given by
    \begin{equation*}
        \widehat{\tau}_{TMLE} = \frac{1}{n} \sum_{i = 1}^n \left[ W_i \times \frac{Y_i - Q(1, X_i)}{\widehat{e}(X_i)} - (1 - W_i) \times \frac{Y_i - Q(0, X_i)}{1 - \widehat{e}(X_i)} + Q(1, X_i) - Q(0, X_i) \right].
    \end{equation*}
The standard error is estimated in the same way as for DRE except that $\widehat{\mu}(w, x)$ is replaced with $Q(w, x)$ and $\widehat{\tau}_{DRE}$ is replaced with $\widehat{\tau}_{TMLE}$.

%###############
\subsubsection{Where Deep Learning Can Be Useful}
We can take advantage of the deep learning models to estimate $\widehat{\mu}(1, x), \widehat{\mu}(0, x)$, and $\widehat{e}(x)$ for DRE and TMLE. There are two slightly different approaches here: 

\begin{itemize}
    
    \item We can use the basic form of deep learning (Multilayer Perceptron) with pre-processed covariates as described in Section \ref{subsec:prepro} such that each loan description will correspond to one single high-dim vector.  
    
    \item We can use more advanced deep learning models such as Recurrent Neural Network or its variations directly with the original word embedding vectors. In this approach, each loan description will correspond to a set of high-dim vectors. 
    
\end{itemize}

%% file: whyDL.tex
\section{Deep Learning Techniques}
\label{sec:DLTechniques}

Deep learning is a cutting-edge method within the machine learning literature, which has been demonstrated to outperform other methods when performing prediction tasks, especially when the data is abundant (LeCun et al. \cite{LBH2015}, Schmidhuber \cite{S2015}). 

In this paper, we use textual data from Kiva, which after being transformed to numerical vectors, will be high-dimensional; these data fit deep learning very well. In this section, we describe in details the deep learning techniques used in this paper.  

\subsection{Preprocessing}
\label{subsec:preprocess}
To process the text data, we use GloVe (Pennington et al. \cite{PSM2014}). The main idea behind GloVe is to find a vector representation for each word in a way that makes sense both semantically (meaning) and syntactically (grammar). A good vector representation of words, called embeddings, should be able to capture such a relation as the famous one below:
\begin{equation*}
    king - man + woman \approx queen.
\end{equation*}
From word representation vectors, we can proceed in two ways:
\begin{itemize}

    \item Use bag-of-word embedding method to create a representative vector for each text paragraph (i.e., loan description). We call these loan vectors. They are simply the averages of all word vectors in the corresponding loan descriptions. These loan vectors will then be used to feed in the baseline model or Multilayer Perceptron model.
    
    \item Keep word vectors as they are and feed them in Recurrent Neural Network models.  

\end{itemize}
Note that we will not retrain the GLoVe model. Instead, we use the pre-trained GloVe vectors on Wikipedia data. Since the point of word vector representations is to create a generally meaningful mapping between words and numerical vectors, these pre-trained vectors would work well in our specific problem. 

\subsection{Multilayer Perceptron}
\label{subsec:MP}

\begin{figure}[h]
    \centering
    \includegraphics[width = 0.8\textwidth]{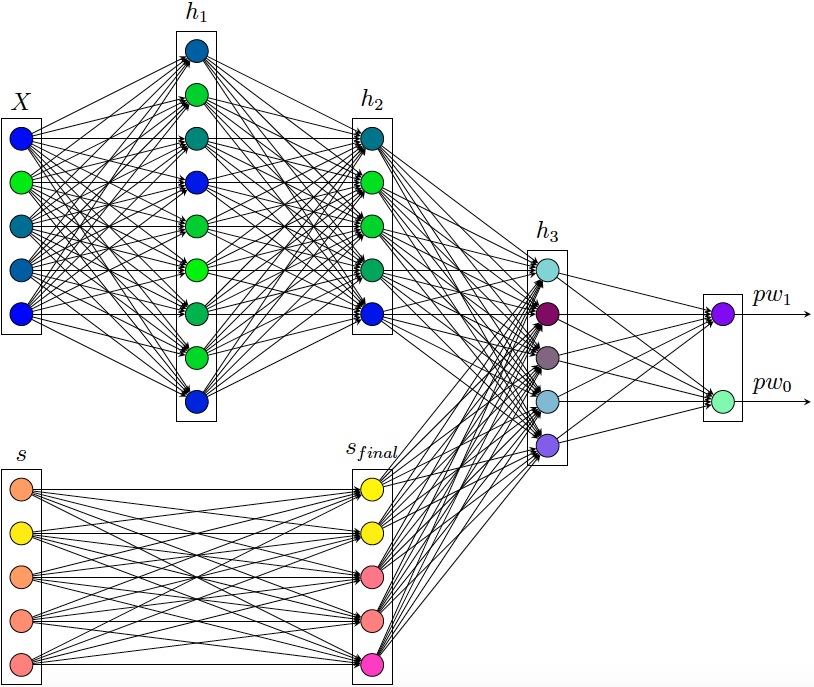}
    \caption{Multilayer Perceptron Model.}
    \label{fig:MLP}
\end{figure}

Assume we want to estimate the propensity score with Multilayer Perceptron (MLP). The model is visualized in Figure \ref{fig:MLP}. 

A deep learning or a neural network model consists of layers of neurons connecting to one another. The model starts with the input layer of dimension $d$, where $d$ is the dimension of the loan vectors. In this paper, we choose $d = 100$. Separately, we create another layer, called $s$ layer, of other covariates (features) of dim $T = 17$.

The first hidden layer has size $n_1$, and each neuron from the input layer connects with every neuron in this layer. The second hidden layer has size $n_2$, and each neuron from the first layer connects with every neuron in this layer. Separately, we create another layer, called $s_{final}$ layer, of dim $n_T$, and each neuron from the $s$ layer connects with every neuron in this layer. We concatenate layer $s_{final}$ to the second layer to increase its size to $(n_2 + n_T)$. The reasoning for concatenating spending data in this stage instead of the input layer is that this extra data is much more important than the the high dimensional loan vector coordinates. Hence, incorporating this extra data with this layer means that we incorporate only the highly influential factors together.
    
The third hidden layer has size $n_3$, and all neurons in the second layer connect will all neurons in this layer. The output layer has dim $2$ whose neurons connect with all neurons in the third hidden layer. For regularization, we use ridge or $L_2$ and dropout. Dropout is a popular regularization technique in deep learning, that randomly drops neurons by setting some vector components to zero.

Mathematically speaking, we need to estimate the parameters $A_1 \in \mathbb{R}^{n_1 \times d}$, $A_2 \in \mathbb{R}^{n_2 \times n1}$, $A_s \in \mathbb{R}^{n_T \times T}$, $A_3 \in \mathbb{R}^{n_3 \times (n_2 + n_T)}$, $A_4 \in \mathbb{R}^{2 \times n_3}$, $b_1 \in \mathbb{R}^{n_1}$, $b_2 \in \mathbb{R}^{n_2}$, $b_s \in \mathbb{R}^{T}$, $b_3 \in \mathbb{R}^{n_3}$, and $b_4 \in \mathbb{R}^2$. The relations are as follows.

\begin{itemize}

    \item $h_1 = ReLU(A_1 X + b_1)$;
    
    \item $h_2 = ReLU(A_2 h_1 + b_2)$;
    
    \item $s_{final} = ReLU(A_s s + b_s)$. Let $H_2 = [h_2, s_{final}]^\prime$ where $s \in \mathbb{R}^T$;
    
    \item $h_3 = ReLU(A_3 H_2 + b_3)$;
    
    \item $out = softmax(A_4 h_3 + b_4)$. 
    
\end{itemize}
Here, we use the Rectified Linear Unit (ReLU) as the ``activation function" where 
\begin{equation*}
    ReLU(x) = \max(0, x).
\end{equation*}  

To estimate the parameters of this model, we minimize the cross-entropy loss function on the $out$ layer. To solve this optimization problem, we use a variation of the gradient descent method with the main factor called back-propagation. A more detailed description of the MLP model and its estimation can be found in Pham \cite{P2016}. 

As we can see, MLP model is much more complex than the simple Logistic Regression one. This complexity creates flexibility in the model, allowing it to capture the relations that cannot be captured in Logistic model. \\

\noindent \emph{Remark:} For (treated and control) outcome model estimations, we use the same neural structure except for two main differences. First, in the output layer we use ReLU instead of softmax function because the outcome now is a continuous value. Second, since the outcome now is a lot more variable we use a ``deeper network'' with two more layers after Layer $h_3$ before reaching the output.

\subsection{Recurrent Neural Network}
\label{subsec:RNN}

\begin{figure}[h]
    \centering
    \includegraphics[width = 1.0\textwidth]{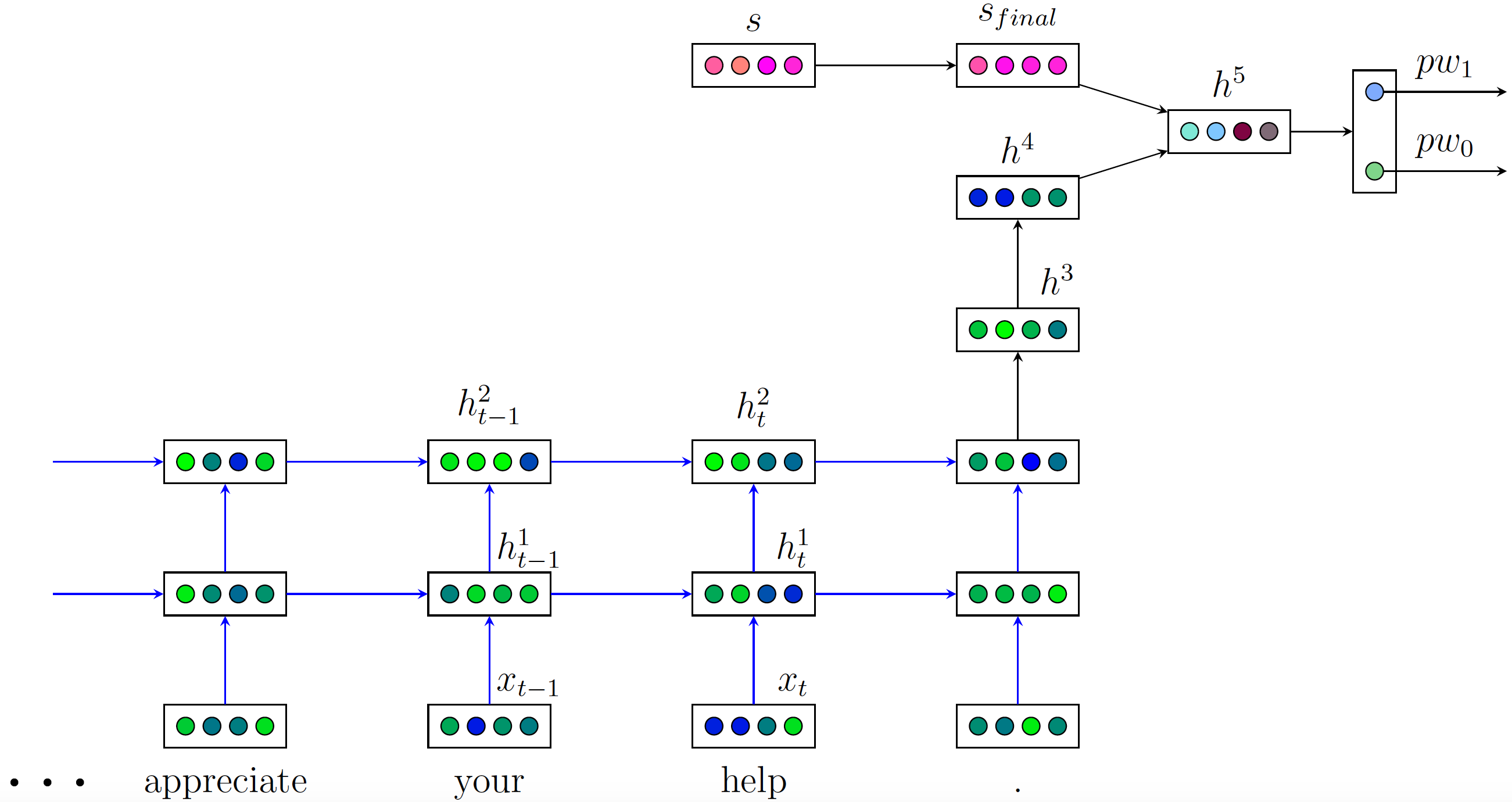}
    \caption{Deep LSTM Model.}
    \label{fig:DeepLSTM}
\end{figure}

We use an advanced version of Recurrent Neural Network (RNN) called Deep RNN with Long Short Term Memory (LSTM) cells, or shortly as Deep LSTM. Assume we want to estimate the propensity score with Deep LSTM. We visualize it in Figure \ref{fig:DeepLSTM}.

The arrows in Figure \ref{fig:DeepLSTM} indicate that the inputs at the arrow tails are used to create the outputs at the arrow heads. There are two important differences from the MLP model:
\begin{itemize}

    \item First, the input now is a variable-size sequence of $d$-dim vectors. 
    
    \item The connection between $s$ and $s_{final}$ is the same as that in the MLP model, and the same is true for the connection between $h^5$ to output layer, $s_{final}$ to $h^5$, $h^4$ to $h^5$, $h^3$ to $h^4$, and the output of the recurrent network to $h^3$. However, the connections in the recurrent networks are much more complex. Specifically, each $h_t^1$ is connected with $h_{t - 1}^1$ and $x_t$ through a necessarily complicated relation called LSTM. Similarly, each $h_t^2$ is connected with $h_{t - 1}^2$ and $h_t^1$ through LSTM. The detail about LSTM is provided in Hochreiter and Schmidhuber \cite{HS1997}. 
    
\end{itemize}

However, this model is similar to the MLP one described above if we consider the output of the recurrent network as the text-data input to feed in other full layers in the same way as in the MLP model. 

Intuitively, this model is designed to capture the sequential relations between pairs of continuous words in the loan description data before outputting an output to feed in other full layers ($h^3, h^4, h^5$) to get the final output. This model is different from MLP model described earlier in one essential aspect: the way we process loan description data. In MLP model, we take the average of all word embedding vectors and therefore neglecting any sequential relations. In Deep LSTM model, however, such relations are kept intact. The deep model is used in stead of the simple model to capture the finer, smoother meaning of the loan description. 

The reader may wonder why we do not keep the relations in the recurrent network the same way as we keep between $s$ and $s_{final}$ and such; this will create a vanilla Recurrent Neural Network. The reason lies exactly in how we estimate this complex model. Similar to estimating MLP model, we need to use some sort of back-propagation techniques to estimate this model. If we use simple relations as those in MLP model here, then we will suffer from the so-called vanishing or exploding gradient issue. We can think about this simply as follows: if we multiply many terms less than one together, then we would get a term close to zero; if we multiply many terms greater than one together, then we would get a very large term; and either case would hinder the estimating process. The many terms appear in this model only, and not in the MLP model, because only this model contains long sequences. The multiplication appears when we use chain rules in calculating gradients for back-propagation in estimating the model.  

The LSTM model fixes the vanishing and exploding gradient problem by utilizing a clever technique that chooses which terms or parts of terms to use in doing multiplications. This way, we can avoid using too small or too large terms. \\

\noindent \emph{Remark:} To estimate the (treated and control) outcome model, we make three changes to the Deep LSTM model used for propensity score estimation. First, we replace the last activation function with ReLU in the same way as in MLP model. Second, we use a deeper network with two more layers after Layer $h^5$ before reaching the output. Last, we incorporate an ``attention weight'' to the output of the recurrent network. In the Deep LSTM model used for propensity score estimation, we use only the last output of the recurrent network. In Deep LSTM with attention model, we use an attention weight to take the weighted average of all the outputs of the recurrent network; these outputs belong to Layer $h^2$ in the model. The attention weight is estimated simultaneously as we estimate the model. We can think about the attention weight as follows: each output in Layer $h^2$ corresponds to a word in the input layer so each output works as a version of a word embedding vector; the attention weight allows the model to take a weighed average, in contrast to the simple average, of these embedding vectors. 

%% file: results.tex
\section{Results}
\label{sec:results}

\subsection{Covariate Relatedness Check}
\label{subsec:covRelCheck}

In Section \ref{sec:pre_analysis}, we see the statistically significant effect of loan amount on outcome $Y$ of our interest. In this subsection, we run a linear regression of $Y$ on loan embedding vector for treated and control groups to see if the text data has statistically significant impact on outcome $Y$.  

Our result shows that among $100$ embedding covariates that are created from loan text data, 83 covariates have statistically significant estimated coefficients on the treated outcome model, and 92 on the control one. This means the text data correlates well with the outcome variable $Y$.

Now running a linear regression of $W$ on loan embedding vector and covariates, we obtain that all but four covariates are statistically significant. Meanwhile, running a linear regression of $Y$ on $W$, loan embedding vector, and other covariates, we obtain that all but seven covariates are statistically significant. This implies two things. First, the text data is important to explain for both the treatment and the outcome. Second, both the p-score and outcome models are dense. 

\subsection{Estimation Model Comparison}
\label{subsec:estModCom}

We first compare different models in terms of propensity score and outcome (both treated and control) estimations. We consider two baseline models: Regularized Logistic/Linear Regression (RLR) and Random Forest (RF), and compare them with two advanced deep learning models: MLP and Deep LSTM. The results are reported in Table \ref{tab:results}.

\begin{table}[h]
\caption{Result Comparison on the Test Set. For p-score estimation, we use $F_1$ score and accuracy ($F_1$ score and accuracy are both in $[0, 1]$; the higher the better); for outcome estimations, we use RMSE (the lower the better).}
\label{tab:results}
\begin{center}
\begin{tabular}{@{}l C{1.5cm} C{1.5cm} C{2cm} C{2cm}@{}}
\toprule
 & \multicolumn{2}{c}{p-score} & Treated & Control \\
\cmidrule(r){2-3} 
\multicolumn{1}{l|}{Method} & ($F_1$) & (acc) & (RMSE) & (RMSE) \\
\midrule
\textbf{w/o text} & & & \\
\multicolumn{1}{l|}{\texttt{RLR}} &  0.41 & 88.2\% & 8.94 & 9.30 \\
\multicolumn{1}{l|}{\texttt{RF}} & 0.56 & 90.2\% & 8.82 & 8.81 \\ \hline
\textbf{with text} & & & \\
\multicolumn{1}{l|}{\texttt{RLR}} & 0.80 & 94.6\% & 8.63 & 9.15 \\ 
\multicolumn{1}{l|}{\texttt{RF}} & 0.82 & 95.5\%  & 7.91 & 8.32 \\
\multicolumn{1}{l|}{\texttt{MLP}} & 0.95 & $98.6\%$  & 7.27 & 7.76 \\
\multicolumn{1}{l|}{\texttt{\textcolor{blue}{Deep LSTM}}} & \textbf{\textcolor{blue}{0.98}} & \textbf{\textcolor{blue}{99.3\%}} & \textbf{\textcolor{blue}{7.24}} & \textbf{\textcolor{blue}{7.70}} \\
\bottomrule
\end{tabular}
\end{center} 
\end{table}

We note that in estimating these models, we use a data-driven approach. That is, we randomly partition the dataset into training, validation, and test set. We use training and validation sets (possibly with cross-validation) to estimate the models. We use a hold-out test set to compare the models' performances.

Getting good performance on a hold-out test set is important. Such a good performance means the model can generalize well to unseen data. Since we do not know all the data for the whole population, a well generalizable model will give better estimates and therefore, the (population) ATE estimate will be more accurate.  

As we can see in Table \ref{tab:results}, the deep learning models outperform other methods. Particularly, Deep LSTM does very well in estimating the propensity scores. In outcome model estimations, MLP and Deep LSTM outperform RLR and RF too, though not by a large margin. This might be because there are some extreme outliers in the test set that no model can account for in the estimating process. 

We also note the role of text data here. Without text data, all model estimations are worse than themselves with it. The text data appears to be especially important to the p-score model as it boosts the $F_1$ score of both RLR and RF a great deal. 

\subsection{Comparison of Average Treatment Effect Estimators}
\label{subsec:ATEestCom}

Now, we compare different methods in terms of ATE estimation. 
\begin{table}[h]
\caption{Summary of ATE Estimation of Different Methods.}
\label{tab:ateEst}
\begin{center}
\begin{tabular}{@{}l|C{2cm} C{2cm}@{}}
\toprule
\multicolumn{1}{l|}{Method}     & ATE   & std \\
\midrule
\textbf{w/o text} & & \\
\multicolumn{1}{l|}{\texttt{Naive}} & 0.17 & 0.027 \\
\multicolumn{1}{l|}{\texttt{Baseline}} & 2.28 & 0.026 \\ \cmidrule{2-3}
\multicolumn{1}{l|}{\texttt{DSE}} &  -0.70 & 0.026 \\ \cmidrule{2-3}
\multicolumn{1}{l|}{\texttt{DRE (RLR)}} &  -0.61 & 1.623 \\
\multicolumn{1}{l|}{\texttt{DRE (RF)}}  & -2.62 & 0.475 \\ \cmidrule{2-3} 
\multicolumn{1}{l|}{\texttt{TMLE (RLR)}} & -1.03 & 1.690 \\ 
\multicolumn{1}{l|}{\texttt{TMLE (RF)}} & -4.38  & 0.644 \\ 
\hline
\textbf{with text} & & \\
\multicolumn{1}{l|}{\texttt{Baseline}} & 2.87 & 0.026 \\ \cmidrule{2-3}
\multicolumn{1}{l|}{\texttt{DSE}} & -0.52  & 0.025 \\ \cmidrule{2-3}
\multicolumn{1}{l|}{\texttt{DRE (RLR)}} & -0.19  &  1.623 \\
\multicolumn{1}{l|}{\texttt{DRE (RF)}} & -1.26 & 0.472 \\
\multicolumn{1}{l|}{\texttt{DRE (MLP)}} & -2.93 & 0.090 \\
\multicolumn{1}{l|}{\texttt{DRE (\textcolor{blue}{Deep LSTM})}} & \textbf{\textcolor{blue}{-3.30}} & \textbf{\textcolor{blue}{0.167}} \\ \cmidrule{2-3}
\multicolumn{1}{l|}{\texttt{TMLE (RLR)}} & -1.00 & 0.990 \\
\multicolumn{1}{l|}{\texttt{TMLE (RF)}} & -12.60 & 0.150 \\
\multicolumn{1}{l|}{\texttt{TMLE (MLP)}} & -2.78 & 0.091 \\
\multicolumn{1}{l|}{\texttt{TMLE (\textcolor{blue}{Deep LSTM})}} & \textbf{\textcolor{blue}{-3.29}} & \textbf{\textcolor{blue}{0.167}} \\
\bottomrule
\end{tabular}
\end{center} 
\end{table}

As can be seen in Table \ref{tab:ateEst}, the naive estimator would conclude that forming group loans has no effect on funding time or it might increase the funding time. The baseline model gives statistically significant positive estimate of ATE; this means that this estimator concludes forming group loans increase the funding time. These results are opposite to those obtained by using advanced deep learning techniques. 

DREs with deep learning give almost identical estimates on the ATE. Surprisingly, DRE with RF without using text data gives a close ATE estimate too. However, its standard deviation is much higher than that of the deep learning approach. So a close estimate of DRE with RF to that of deep learning approach could be explained by randomness. DSE also gives negative effect estimate with and without text data though the magnitude is small.
 
TMLEs with deep learning give quite different estimates, though all indicate a significantly negative effect. The estimate by TMLE with RF without text data is close but it could be due of randomness. 

Note that both DRE and TMLE possess the double robustness property. The Deep LSTM used to estimate the propensity score in these two models gives us an almost perfect estimate. In other words, the propensity score model is mostly correctly specified. Double robustness would imply that DRE and TMLE are consistent estimators of the ATE. Hence, we would rely on the DREs and TMLEs with MLP and especially Deep LSTM estimation methods. Note that we use only observation with estimated propensity scores in [0.01, 0.99] interval because extremely small propensity scores would behave very weirdly. We conclude that forming group loans shortens the funding time by about $3.3$ days on average.

%% file: conclusion.tex
\section{Conclusion}
\label{sec:conclusion}
In this paper, we have shown the important role of deep learning in dealing with unstructured text data to answer causal questions. Combining deep learning techniques with causal inference framework could bring more precise estimation results of causal effects. Specifically, we use deep learning to estimate the infinite dimensional components of the Doubly Robust Estimator and Targeted Maximum Likelihood Estimator while using influence curves to estimate the treatment effects of interest. The results show that deep learning models outperform other approaches in estimating these components, i.e. propensity score and outcome models.

We find that on average, forming group loans has a significant treatment effect on funding time. In particular, pooling multiple loans into a group takes about $3.3$ days fewer to get funded than posting them individually on Kiva. Hence, it is advisable for field partners to pool the borrowers and introduce them as a group. On lender's side, it is also less riskier to diversify their loaning portfolio.

For future work, we aim at extending the investigation to estimating average effects in important, popular sub-groups of borrowers corresponding to loan categories such as loans for agriculture, loans for food, and retail loans. We expect that the treatment effects across different sub-groups will be different.

%% file: reference.tex
\newpage
% ------------- & ------------- % -------------- & ------------- % ------------- & -------------- %
%%%%%%%%%%%%%%%%%%%%%%%%%%%%%%%%%
%%%%%%%%%%%%%%%%%%%%%%%%%%%%%%%%%
%%%%%%%%%%%%%%%%%%%%%%%%%%%%%%%%%

%% file: appendix.tex
\begin{appendices}

%%%%%%%%%%%%%%%%%%%
%%%%%%%%%%%%%%%%%%%
%%%%%%%%%%%%%%%%%%%
\section{Raw Data Description}

\subsection{Raw Data Used}

We use data retrieved from the Kiva website from Jan $1^{st}$, 2006 to May $10^{th}$, 2016, which are stored in \textit{json} format. There are $2240$ such files. We use only loan data in this project. 

\subsection{Raw Data Attributes}

The data is stored in json format in a dictionary type. The original attributes (which may themselves be dictionaries) are: \texttt{name}, \texttt{payments}, \texttt{journal\char`_totals}, \texttt{image}, \texttt{themes}, \texttt{tags}, \texttt{translator}, \texttt{currency\char`_exchange\char`_loss\char`_amount}, \texttt{basket\char`_amount}, \texttt{video}, \texttt{description}, \texttt{borrowers}, \texttt{posted\char`_date}, \texttt{funded\char`_date}, \texttt{activity}, \texttt{arrears\char`_amount}, \texttt{bonus\char`_credit\char`_eligibility}, \texttt{delinquent}, \texttt{funded\char`_amount}, \texttt{id}, \texttt{location}, \texttt{paid\char`_amount}, \texttt{paid\char`_date}, \texttt{partner\char`_id}, \texttt{planned\char`_expiration\char`_date}, \texttt{use}, \texttt{terms}, \texttt{lender\char`_count}, \texttt{loan\char`_amount}, \texttt{sector}, \texttt{status}.
Most of the attributes are in \textit{string} format; only a few are \textit{numerical} or \textit{categorical}. On a side note, many attributes have blank or identical values.

\subsection{Sample Raw Data}

As we mentioned above, not all attributes have values. We provide here an example for those attributes which have numerical or non-numerical values. 

\texttt{activity}: Higher education costs

\texttt{bonus\char`_credit\char`_eligibility}: False

\texttt{borrowers}: [$\{$first$\_$name: Mahesh, gender: M, last$\_$name: '', pictured: True$\}$]

\texttt{description}: $\{$languages: [en], texts: $\{$en: Extra efforts, all-nighters, and sacrifices are always worth it when it come to studies because studies are the only sure way to improve one 2019s quality of life. Mahesh is well aware of this, because he has always counted his goal as becoming a good banker and financial adviser. Now he has reached the last step in completing his degree and entering the working era that any professional lives in. Mahesh is asking for a loan to pay off his PGDBM (Post graduate diploma in banking management) course fees and accomplish his dream. He also hopes to specialize in banking and Finance, and in the long run, to establish a financial advisory services. This enthusiastic young man of only 24 years urges young people to make an effort reaching for their dreams because, just like him, they can get ahead.$\}\}$

\texttt{funded\char`_amount}: 1150

\texttt{funded\char`_date}: 2015-03-24T06:06:28Z

\texttt{id}: 853701

\texttt{image}: {id: 1833527, template$\_$id: 1}

\texttt{journal\char`_totals}: {bulkEntries: 0, entries: 0}

\texttt{lender\char`_count}: 46

\texttt{loan\char`_amount}: 1150

\texttt{location}: $\{$country: India, country$\_$code: IN, geo: $\{$level: town, pairs: 20 77, type: point$\}$, town: Head Office, Bhubaneswar, Odisha$\}$

\texttt{name}: Mahesh

\texttt{partner\char`_id}: 241

\texttt{planned\char`_expiration\char`_date}: 2015-04-17T18:20:05Z

\texttt{posted\char`_date}: 2015-03-18T18:20:05Z

\texttt{sector}: Education

\texttt{status}: funded

\texttt{tags}: [$\{$name: $\#$Schooling$\}$]

\texttt{terms}: $\{$disbursal$\_$amount: 72000, disbursal$\_$currency: INR, disbursal$\_$date: 2015-02-10T08:00:00Z, loan$\_$amount: 1150, local$\_$payments: [ ], loss$\_$liability: $\{$currency$\_$exchange: shared, currency$\_$exchange$\_$coverage$\_$rate: 0.1, nonpayment: lender$\}$, repayment$\_$interval: None, repayment$\_$term: 43, scheduled$\_$payments: [ ]$\}$

\texttt{themes}: [Higher Education]

\texttt{translator}: $\{$image: 1126934, byline: Erin Truax$\}$

\texttt{use}: to pay for his PGDBM course fees.

\section{Data Manipulation}
We process the raw data to create the covariates, treatment, and outcome needed for the estimations of our models.

\begin{enumerate}

    \item There are $330$ loans that are already funded before posting the request on Kiva. We discard these loans from consideration.
     
    \item We discard unnecessary covariates whose values are mostly None or identical. 
    
    \item We discard covariates whose values are hard to process such as \texttt{location} or seemingly unimportant such as \texttt{partner\char`_id}, \texttt{image}, \texttt{journal\char`_totals}, and \texttt{id}. 
    
    \item We discard covariates which happen after the funding decision such as \texttt{lender\char`_count} and \texttt{payments}. 
     
    \item We create \texttt{description\char`_language} and \texttt{description\char`_texts} separately from \texttt{description}. Then, we discard \texttt{description} as well its shorter versions: \texttt{activity}, \texttt{use}, and \texttt{themes}. We also keep only examples with English descriptions.
    
    \item We discard loan entries in which \texttt{description\char`_texts} is empty or has no actual text value. 
    
    \item We create the treatment variable $W$ from \texttt{borrowers} by simply counting the number of borrowers there. We discard examples in which there is no borrower. If there are multiple borrowers, then $W = 1$; otherwise, $W = 0$.
    
    \item We create the outcome variable $Y$ by taking the difference between \texttt{funded\char`_date} and \texttt{posted\char`_date} and converting this value into day unit. If the project is never funded, then $Y$ is \text{infinity}. In this paper, we consider only funded loans so we discard all examples in which $Y$ is infinity. 
    
    \item We obtain variable \texttt{gender} by taking the majority of the borrowers' genders. Then, we discard \texttt{borrowers}. 
    
    \item We obtain the binary variable \texttt{risker} whose value is $0$ if the partner bears the default risk and $1$ if the lender does.   
    
    \item From the categorical variable \texttt{sector} which consists of $15$ categories, we create $14$ dummy variables corresponding to $14$ categories and discard \texttt{sector}. 
    
\end{enumerate}
After manipulating the data, we obtain $995,911$ loan entries with the text covariate \texttt{description\char`_texts} and other $17$ covariates which includes $14$ sector dummy variables, \texttt{loan\char`_amount}, \texttt{risker}, and \texttt{gender}. Among these $17$ covariates, only \texttt{loan\char`_amount} is non-binary so we normalize it to have zero mean and unit variance.

\section{Deep Learning Models}

\begin{enumerate}

    \item Use GloVec pre-trained word vectors (trained on Wikipedia data) as the fixed representations of the words in our models. We use 100-dimensional pre-trained GloVe vectors.
    
    \item We start with Multilayer Perceptron model and then move to RNN with LSTM cells, specifically Deep LSTM. For Multilayer Perceptron, we take the average of the word vectors in each loan description to obtain the loan vectors. For Deep LSTM, we keep word embeddings vectors at the word level.  
    
    \item For regularization, we use both $L_2$ loss and dropout.  
    
    \item All model estimations (i.e., trainings) are done in Tensorflow with GPUs. 

\end{enumerate}

\end{appendices}